\documentclass[lettersize,journal]{IEEEtran}

\usepackage{times}  % DO NOT CHANGE THIS
\usepackage{helvet}  % DO NOT CHANGE THIS
\usepackage{courier}  % DO NOT CHANGE THIS

\usepackage{graphicx}
\usepackage{booktabs}

\usepackage[caption=false]{subfig}
\usepackage{bm}
\usepackage{amsmath}
\usepackage{amsfonts}
\usepackage{cite}

\usepackage{xcolor}
\usepackage[pagebackref=true,breaklinks=true,colorlinks,bookmarks=false]{hyperref}

\begin{document}
\title{Multimodal Neurodegenerative Disease Subtyping Explained by ChatGPT}
%\author{First Author\inst{1}\orcidID{0000-1111-2222-3333} \and Second Author\inst{2,3}\orcidID{1111-2222-3333-4444} \and Third Author\inst{3}\orcidID{2222--3333-4444-5555}}
%

\author{Diego Machado Reyes, Hanqing Chao, Juergen Hahn, Li Shen~\IEEEmembership{Senior~Member,~IEEE}, 
Pingkun~Yan,~\IEEEmembership{Senior~Member,~IEEE} and for the Alzheimer's Disease Neuroimaging Initiative\*
\thanks{Manuscript received on xxx, 2023.}%
\thanks{D. Machado Reyes, H. Chao, J. Hahn, and P. Yan are with the Department of Biomedical Engineering and the Center for Biotechnology and Interdisciplinary Studies, Rensselaer Polytechnic Institute, Troy, NY, USA 12180;}%
\thanks{L. Shen is with the Department of Biostatistics, Epidemiology and Informatics, Perelman School of Medicine at the University of Pennsylvania, Philadelphia, USA}
\thanks{This work was supported in part by the NSF CAREER award 2046708, NSF IIS 1837964, and NIH grants U01 AG066833, U01 AG068057, and RF1 AG063481.}%
\thanks{\*Data used in preparation of this article were obtained from the Alzheimer's Disease Neuroimaging Initiative (ADNI) database (adni.loni.usc.edu). As such, the investigators within the ADNI contributed to the design and implementation of ADNI and/or provided data but did not participate in analysis or writing of this report. A complete listing of ADNI investigators can be found at: \url{http://adni.loni.usc.edu/wp-content/uploads/how_to_apply/ADNI_Acknowledgement_List.pdf}}
}
% \institute{**********}
% \author{Anonymous}
%

%\institute{Princeton University, Princeton NJ 08544, USA \and Springer Heidelberg, Tiergartenstr. 17, 69121 Heidelberg, Germany \email{lncs@springer.com}\\ \url{http://www.springer.com/gp/computer-science/lncs} \and ABC Institute, Rupert-Karls-University Heidelberg, Heidelberg, Germany\\ \email{\{abc,lncs\}@uni-heidelberg.de}}
%
\maketitle              % typeset the header of the contribution
\begin{abstract}
% Background\\
% Purpose\\
% Method\\
% Result\\
Alzheimer's disease (AD) is the most prevalent neurodegenerative disease; yet its currently available treatments are limited to stopping disease progression. Moreover, effectiveness of these treatments is not guaranteed due to the heterogenetiy of the disease. Therefore, it is essential to be able to identify the disease subtypes at a very early stage. Current data driven approaches are able to classify the subtypes at later stages of AD or related disorders, but struggle when predicting at the asymptomatic or prodromal stage. Moreover, most existing models either lack explainability behind the classification or only use a single modality for the assessment, limiting scope of its analysis. Thus, we propose a multimodal framework that uses early-stage indicators such as imaging, genetics and clinical assessments to classify AD patients into subtypes at early stages. Similarly, we build prompts and use large language models, such as ChatGPT, to interpret the findings of our model. In our framework, we propose a tri-modal co-attention mechanism (Tri-COAT) to explicitly learn the cross-modal feature associations. Our proposed model outperforms baseline models and provides insight into key cross-modal feature associations supported by known biological mechanisms.     
\end{abstract}

\section{Introduction}

\IEEEPARstart{A}{lzheimer}'s disease (AD) is the most prevalent neurodegenerative disorder, which affects over 6.5 million people in the US alone and its rate is expected to keep increasing \cite{noauthor_2022_2022}. Current therapies for AD are limited to symptom management and promising drugs that may slow AD progression; thus, it is vital to diagnose neurodegenerative disease early. However, early diagnosis of AD is very challenging as the characteristic symptoms used in clinical assessments appear only at intermediate or later stages and can vary among patients due to the disease heterogeneity. Therefore, it is essential to develop methods that can characterize the factors driving disease development and identify the specific subtype for individual patients at an early stage.

% Merged from related works
AD is traditionally diagnosed based on characteristic cognitive decline and behavioral deficits that do not become apparent until intermediate to late stages of the disease.
% Moreover, clinical assessment of subtypes focuses on two approaches , either distinguishing patients based on the age onset  or based on the predominant aggregation of tau in specific regions. Patients are usually classified as typical or atypical AD~\cite{vogel_subtypes_2022}. Similarly, studies have pushed towards three biologically different subtypes based on the distribution of tau-related pathology~\cite{ferreira_biological_2020}. On the other hand,
Data driven approaches focus on classifying patients into subtypes based on the disease progression and mild cognitive impairment (MCI, a prodromal stage of AD) to AD conversion~\cite{mitelpunkt_novel_2020,badhwar_multiomics_2020,marti-juan_revealing_2019,feng_deep_2022}. 
%
%they are constrained to aggregated feature representations. 
% On the other hand, machine learning approaches are limited to reporting feature importance individually such as done in unified methods as in LIME~\cite{ribeiro_why_2016} and SHAP~\cite{lundberg_unified_2017}. However, biomarkers work together in biological pathways towards disease development. Similarly, traditional deep learning models are limited to post-hoc feature importance analysis and lack explainabiltiy regarding the feature associations~\cite{guidotti_survey_2018} .
%
Current methods for AD and related disorders subtyping have mainly focused on using the longitudinal data from clinical assessments for unsupervised learning~\cite{emon_clustering_2020,wen_multi-scale_2022,poulakis_multi-cohort_2022}. For clustering approaches such as in~\cite{marti-juan_revealing_2019,mitelpunkt_novel_2020,emon_clustering_2020,poulakis_multi-cohort_2022}, the authors subtype AD patients using single modalities at baseline, such as blood markers~\cite{marti-juan_revealing_2019}, genomic data~\cite{emon_clustering_2020}, or imaging derived traits from longitudinal measurements~\cite{poulakis_multi-cohort_2022}. These single modality approaches at baseline data have shown the potential of early stage indicators for AD subtyping. However, most of them miss to show how they relate biologically to AD development or use multiple time points, which hinders the ability to diagnose patients at early stages after AD onset. 

Deep learning models~\cite{feng_deep_2022} have effectively identified subgroups using multimodal imaging data and correlation-based approaches which allow for increased explainability of the model feature relationships. 
Nevertheless, correlation-based approaches treat the clustering goal indirectly. Works on related disorders have shown the relevance of multimodal approaches~\cite{dadu_identification_2022} using non-negative matrix factorization and Gaussian mixture models, or employed autoencoders and long short-term memory networks (LSTM) to learn deep embeddings of the disease progression~\cite{su_integrative_2022}. However, this requires longitudinal data involving several time points. While several clustering and deep learning-based approaches have high accuracy when having multiple time points, the performance decreases significantly given only baseline data. This is driven by the subtle expression of the symptoms tracked in the clinical assessments limiting the scope of the models. Early stage indicators such as imaging traits or genomic risk factors are rarely used, and are simply aggregated to clinical assessments as additional indicators.
Therefore, it is essential to target early stage indicators such as genetics, imaging and cognitive assessments.

Multimodal deep learning approaches can combine different modalities and provide a much more informed picture of the disease drivers and aid in the disease subtyping~\cite{nguyen_multiview_2020}. However, it is not a trivial task to identify the relevant features across modalities and how to fuse them. The rest of this section first reviews the related works on multimodal fusion and then provides an overview of our proposed method.

\begin{figure*}[t]
	\centering
	\includegraphics[width = \textwidth,trim=0 55 0 105, clip =true]{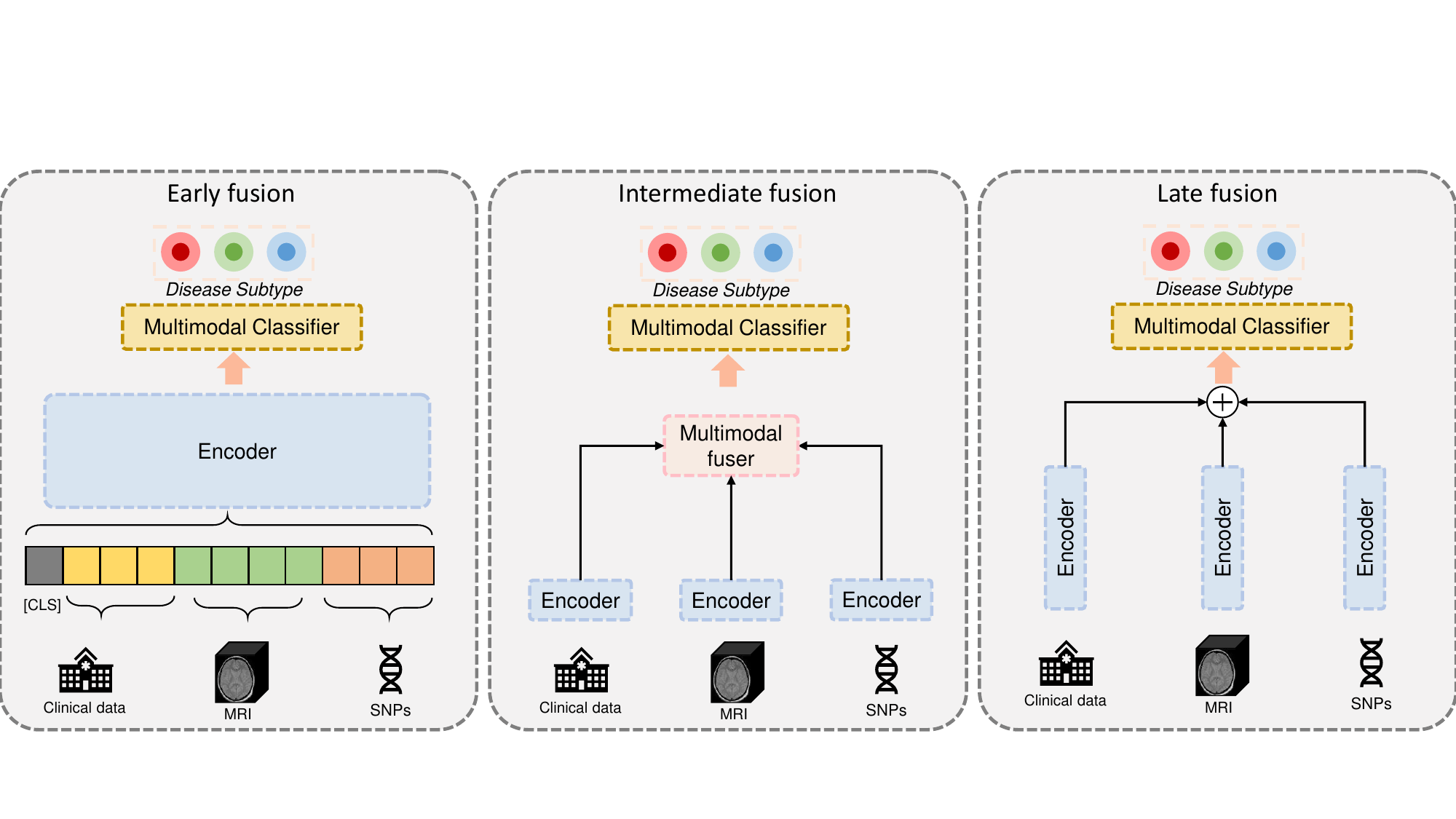}%\hfill
	\caption{The three main multimodal fusion strategies, early, intermediate and late fusion, for deep learning methods. }
	\label{fig:fusion_methods}
\end{figure*}

\subsection{Multimodal Fusion}

Multimodal fusion, while very promising, poses new challenges. There are multiple ways to fuse the data and stages of encoding where to fuse. The strategy effectiveness varies depending on data modalities and tasks. One of the key factors is the similarity between the modalities. Highly heterogeneous data such as imaging, genetics and clinical data, might not be immediately fused. Their difference in type, signal to noise ratio, and dimensionality makes it be very challenging to combine them without first projecting them into a similar space. The relationship between input and output is just as important to consider when fusing data. For example, clinical assessments reflect the direct impacts of the disease, while genetic data describes the building blocks of cells. The phenotype-related information available in clinical assessments requires considerably less processing than what the genomic data might require to find the connection with the disease.

As illustrated in Fig.~\ref{fig:fusion_methods}, the existing multimodal fusion strategies can be grouped into three main categories~\cite{huang_fusion_2020}, including early, intermediate (also called joint) and late fusion. It is essential to choose the right approach based on the task at hand and the data used. Early stage fusion has shown very promising results in recent vision-language models~\cite{alayrac_flamingo_2022,akbari_vatt_2021}, while late-fusion strategies have traditionally been very effective in aggregating machine learning models decisions. Early and late stage fusion strategies, while effective for certain tasks, are not ideal for AD subtyping using multimodal data. Early stage fusion struggles with dealing with highly heterogeneous and differently biologically-related data such as genetics, imaging and clinical. These require further encoding to first learn highly informative representations for every modality and condense them into a similar latent space. While late fusion strategies can be very effective at aggregating the decisions based on each modality, they cannot learn the feature relationships across modalities severely limiting the usefulness of the model.
%The intermediate fusion approach may address both challenges, first by learning the key patterns related to the subtype for each modality allowing the heterogeneous modalities to be more harmoniously fused, and through the multimodal attention mechanism it allows to characterize the cross-modal feature relationships.
The intermediate fusion approach tackles both challenges by first learning the crucial patterns associated with each modality. On the next stage, it uses the condensed patterns from each modality to learn the cross-modal relationships. This enables a more harmonious fusion of the heterogeneous modalities. 

\begin{figure*}[t]
	\includegraphics[width=\textwidth, trim=0 80 0 60, clip =true]{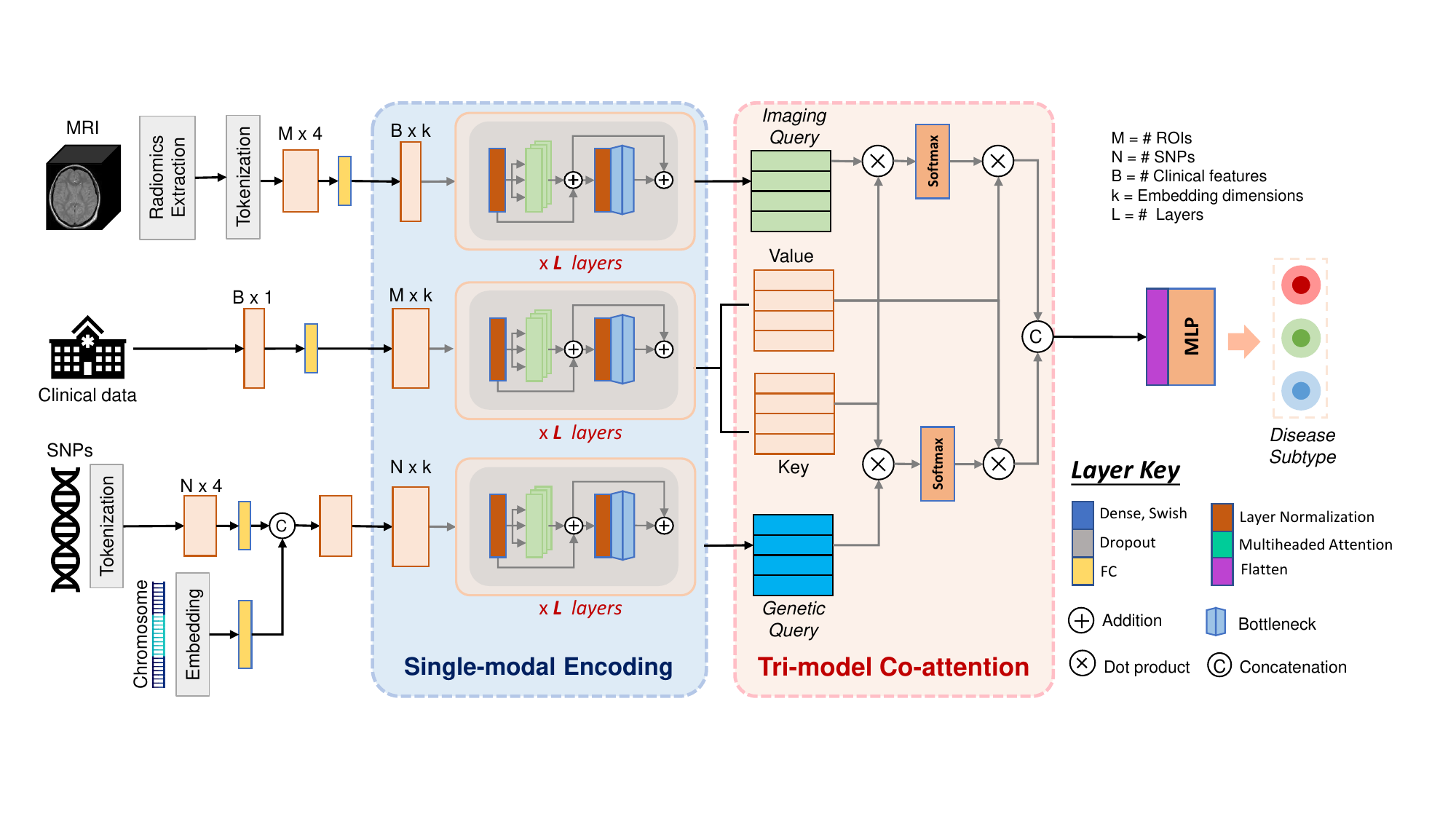}
	\caption{Illustration of the proposed framework for AD subtyping, consisting of two main sections: (a) single modality encoding and (b) tri-modal attention and joint encoding. %The model takes imaging data as its main input and, through the co-attention framework it is able to learn the relationships between the imaging derived features and the genetics and clinical features. 
	}
	\label{fig:framework}
\end{figure*}

%\subsection{Tri-modal Co-Attention}
\subsection{Overview and Contributions}
Despite the potential advantages of multimodal approaches, several technical challenges exist for effectively leveraging the key data from each modality. The high heterogeneity of the data modalities for AD subtyping, and the explicit learning of the cross-modal interactions need to be addressed. Previous approaches in related fields have proposed dual co-attention mechanisms to explicitly learn the cross-modal feature interaction and joint data representations~\cite{chen_multimodal_2021}. While they have shown very promising results, these have yet to be explored for neurodegenerative disease subtyping and its corresponding modalities. Moreover, in AD and related disorders subtyping, we need to fuse three critical but highly heterogeneous modalities, i.e., imaging, genetics and clinical data.
%, which poses the need for a tri-modal co-attention mechanism. 
% 
Therefore, in this paper, we propose a tri-modal co-attention (Tri-COAT) framework that can explicitly learn the cross-modal feature interactions towards the subtype classification task.
While deep learning models for disease assessment promise improved accuracy, they remain limited in the interpretability of their results. This is a major entry barrier for the inclusion of deep learning models in the medical field.

The advent of large language models has offered new areas of opportunity to bridge the gap between deep learning models and their reach to the general public.  It is vital to develop models that physicians can use as tools that provide clear information of what the model predictions mean beyond a probability score. ChatGPT~\cite{openai_gpt-4_2023} allows for very accessible written outputs from complex prompts. By crafting a prompt using the output information from our proposed network, we can use ChatGPT to interpret the model AD assessment. 

Our contributions in this paper are two fold: 
\textbf{Application}: our framework incorporates features from three early stage biomarker modalities and provides a cutting-edge approach to early neurodegenerative disease subtyping. Through ChatGPT, the model results are interpreted into a highly accessible text output.
\textbf{Technique}: our new tri-modal co-attention can efficiently and explicitly learn the interactions between highly heterogeneous modalities, encode the information into a joint representation, and provide explainability to the cross-modal interactions.
Our proposed method, Tri-COAT, achieved state-of-the-art performances on the landmark dataset Alzheimer’s Disease Neuroimaging Initiative (ADNI),  and provided key insights into the biological pathways leading to neurodegenerative disease development.

\section{Method}

Our proposed framework can be divided into two main parts. As seen in Fig.~\ref{fig:framework}, single modality encoders are first built using transformer modules to learn feature representations for each modality.
%single modality encoders are first built for each modality and representations are learned through individual transformer encoders. 
Second, the Tri-COAT mechanism explicitly learns the critical cross-modal feature relationships and uses that to weigh the feature representation. The jointly learned representation goes through a multilayer perceptron (MLP) for disease subtype classification.

\subsection{Single-modality encoding}

Three branches encode each modality individually. Each branch is comprised of a transformer encoder with several transformer layers. Each branch learns representations of a modality to later combine them into a joint representation through the Tri-COAT mechanism. 

\noindent\textbf{Imaging modality.} 
% \py{Pls provide details of radiomics extraction and tokenization. why do you need them?}
The imaging feature encoder branch uses MRI derived quantitative traits as input. These quantiative traits are derived from T1 weighted
MRI scans. The scans are first segmented based on the FreeSurfer atlas
for cross-sectional processing. Next, for each reconstructed regions of interest (ROI) the cortical region, cortical volume, thickness average, thickness standard deviation, and surface area are calculated. Further details are described in section~\ref{sec:dataset}. The imaging traits are then used to build tokens, where each token represents an ROI in the brain and is comprised of four imaging derived traits (Cortical Thickness Average, Cortical Thickness Standard Deviation, Surface Area, Volume from Cortical Parcellation).
Let $\bm{X_{I}} \in \mathbb{R}^{M \times 4}$ represent the imaging input to the proposed model, where $M = 72$ and is the number of ROIs. After, the token dimensions are expanded through a fully-connected layer to match the model dimensions $k$. The imaging tokenization allows to build an initial representation for each ROI rather than each trait, leading to a smaller number of input features, and a more biologically informative and interpretable input.

\noindent\textbf{Genetic modality.}
% \py{Please explain 1) the representation of SNPs and Genes; 2) Tokenization; 3) SNP + Gene concatenation; and 4) Transformer.}
The genotype branch has as input single nucleotide polymorphism (SNP) data. After quality control and preprocessing of the genotype data as described in section~\ref{sec:dataset}, tokens for each SNP are built. Each token is comprised by the allele dosage from the patient, the corresponding odds ratio and rare allele frequency obtained from the most recent AD GWAS study, and whether the SNP is within an intergenic region (regulatory region) as a binary label. Then, the token dimensions are expanded through a fully-connected layer to $k/2$.  Let $\bm{X_{SNP}} \in \mathbb{R}^{N \times k/2}$ represent the genotype input to the model, where $N = 70$ and is the number of SNPs filtered out (see section 3.1 for details). Moreover, based on the chromosome for which each SNP is located in an additional embedding for each SNP can be built. By including the chromosome embedding, location knowledge for each SNP can be incorporated. Using an embedding layer, an embedding for each chromosome can be obtained $\bm{X_{Chr}} \in \mathbb{R}^{N \times k/2}$. Finally, $\bm{X_{SNP}}$ and $\bm{X_{Chr}}$ are concatenated to obtain the final genotype embedding $ \bm{X_G} \in \mathbb{R}^{N \times k}$. Similarly to the imaging data encoding, the genetic tokenization allows to build more informative input structures to the genetic encoder. By providing additional attributes for each SNP beyond the patient mutation status, the model can learn richer patterns of characteristics that relate each SNP.

% \py{The above description also misses why this may be good. Please justify your design details. The same thing for the description below.}
% Moreover, based on the gene for which each SNP is located in or most closely related an additional embedding for each can be built. Therefore, we can consider each SNP as a token and construct its embedding using the features described. Furthermore, each gene can be encoded using an embedding layer and concatenated to the SNP embedding. 
%Alternatively, instead of the genes, the chromosome could be used as the embedding, thus incorporating some location knowledge for each SNP.% 

\noindent\textbf{Clinical modality.} The clinical data is already very closely related to the outcome of interest and contains only few features; therefore, no further extensive tokenization is performed. As there is just one value per clinical assessment, the tokens are directly built with one dimension. Let $\bm{X_{C}} \in \mathbb{R}^{B \times 1}$ represent the clinical input to our model, where $B = 7$ and is the number of clinical features.  Next, the token dimensions are expanded through a fully-connected layer to match the model dimensions $k$. 

\noindent\textbf{Single modality encoders.} After the tokenization of each modality, they are fed into independent transformer encoders with $L$ layers. The full process of the $L$-th layer in our transformer encoder is formulated as: \begin{equation}  
	F'_l = \mathrm{MHA}(\mathrm{LN}(F_{l-1})) + F_{l-1}, F_l  = \mathrm{FF}(\mathrm{LN}(F'_l)) + F'_l, 
\end{equation} 
where $\mathrm{LN}(\cdot)$ is the normalization layer~\cite{ba_layer_2016}, and $\mathrm{MHA}(\cdot)$ is multihead self-attention~\cite{vaswani_attention_2017}.

\subsection{Tri-modal co-attention}

% \py{Use math symbols and equations to help describe the technical details, like your papers from last year. Those symbols should match what we have in Fig. 1.}
After each transformer encoder has learned a new representation for each modality, these are then used to learn the cross-modal feature relationships to guide the co-attention process on the clinical branch. In other words, the imaging and genomic features are employed to modulate the clinical learning process by highlighting the key hidden features that share relationships across modalities. The intuition behind the proposed approach is that as the clinical data is most closely related to the disease phenotype, this branch will carry most of the necessary information to classify the patients. Nevertheless, the imaging and genomic data provide also valuable information. The idea is analogous to the clinical data being the subject and verb in a sentence while the imaging and genomic data are the adjectives and adverbs. These two elements enrich the representation of the patient health status, analogous to enriching a sentence for a fuller meaning. 

% \py{Please justify this design. Why this may be good?}
Let $\bm{X_{Emb}} \in \mathbb{R}^{\{M,N,B\} \times k}$ represent the learned representation of a given single-modality encoder. These become then query  matrices for the genetics $\bm{Q_{G}}$ and imaging $\bm{Q_{I}}$ data, and key $\bm{K_{C}}$, value $\bm{V_{C}}$ matrices for the clinical data. Following an attention mechanism structure, the co-attention between two modalities is computed as: 
\begin{equation}
\label{equation1}
    \mathrm{CoAttn}(\{G,I\}, C) = \mathrm{softmax} \left( \frac{{Q_{\{G,I\}}} K_C^{T}}{\sqrt{d_k}}\right) V_C^T.
\end{equation}
Next, the resulting co-attention filtered value matrices are then concatenated to obtain a final joint representation. This joint representation is then flattened and used to classify the patients into the clusters through an MLP.

% \subsection{Subtype definition}
% \label{sec:subtype def}

% The labels for the AD subtyping were calculated based on clustering of the cognitive decline over the recorded time points. The Mini-Mental State Examination (MMSE) score at each visit (basline, 6 months, 12 months, and 24 months) is used to determine the cognitive decline for each patient. As each patient may have a different starting level at baseline, the baseline measurement is subtracted from each of the following time points; thus all patients start at 0. Then, using k-means clustering, using $k = 3$, a slow, intermediate and fast cognitive decline groups are defined.

\begin{table}[t]
	\centering
	\caption{Subject data distribution reported as mean $\pm$ standard deviation for MMSE and age, and counts per category for number of participants and gender.} 
	\label{tab:demographics}
	\begin{tabular}{l|c|c|c}
		\toprule
		& \textbf{Slow} & \textbf{Intermediate} & \textbf{Fast}\\\midrule
		Participants  & 177 & 302 & 15 \\
		MMSE (BL)   &   27.35 $\pm$ 2.51    &   27.66 $\pm$ 1.86    &   24.93 $\pm$ 3.55 \\
		MMSE (m24)  &   28.15$\pm$ 2.15     &   23.86 $\pm$ 3.68    &   15.9 $\pm$ 4.84 \\
		Age         &   73.26 $\pm$ 7.82    &   72.44 $\pm$ 7.55    &   71.22 $\pm$ 3.922  \\
		Gender      &   M: 102 F: 75        &   M: 185 F: 117       &   M: 9 F: 6 \\
		
		\bottomrule
	\end{tabular}
\end{table}

\begin{figure}[t]
	\centering
	\subfloat[]{\includegraphics[width = 0.5\textwidth,trim=0 20 40 25, clip =true]{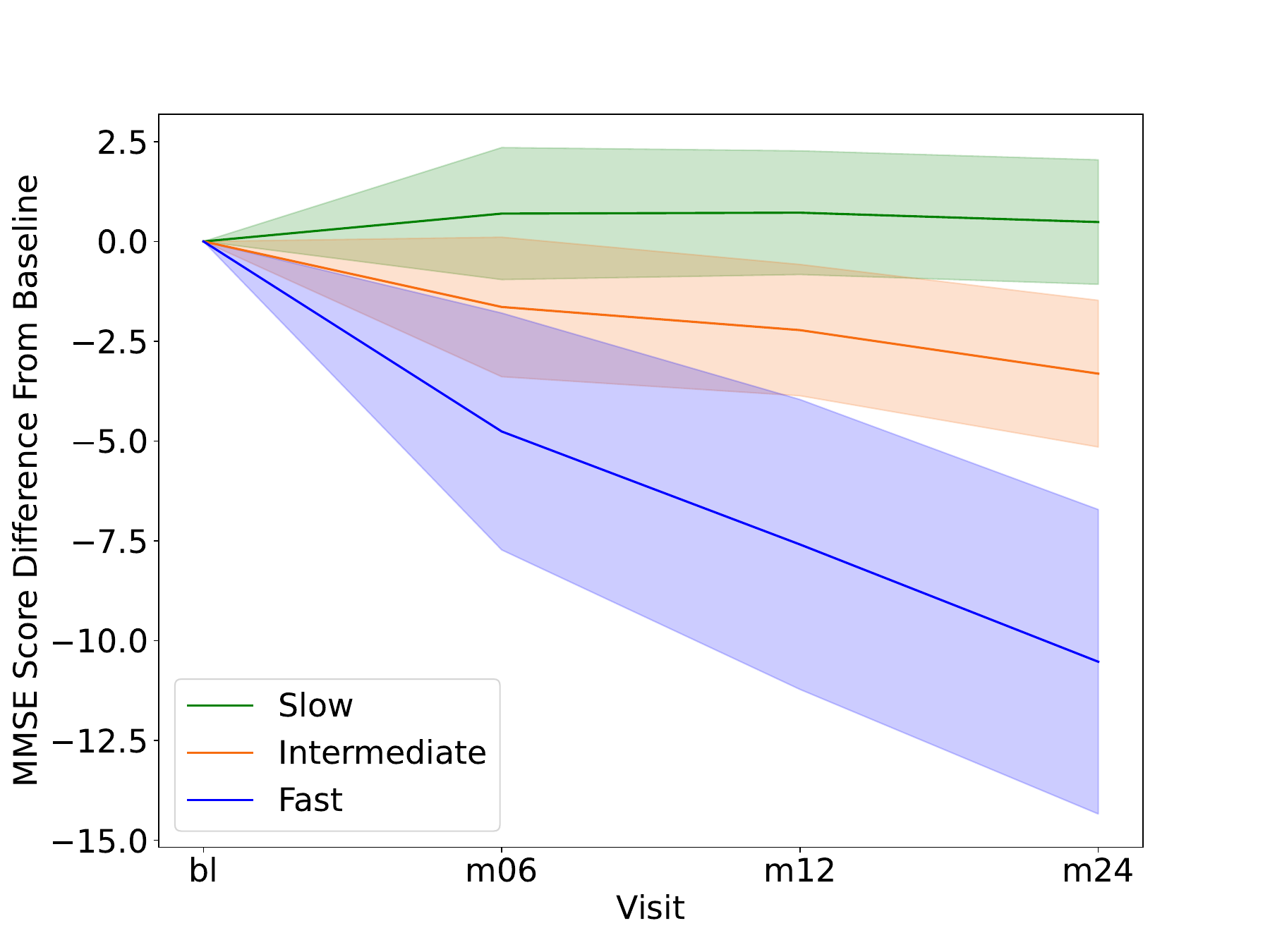}} \\
	\subfloat[]{\includegraphics[width = 0.5\textwidth, trim=25 20 40 25, clip =true]{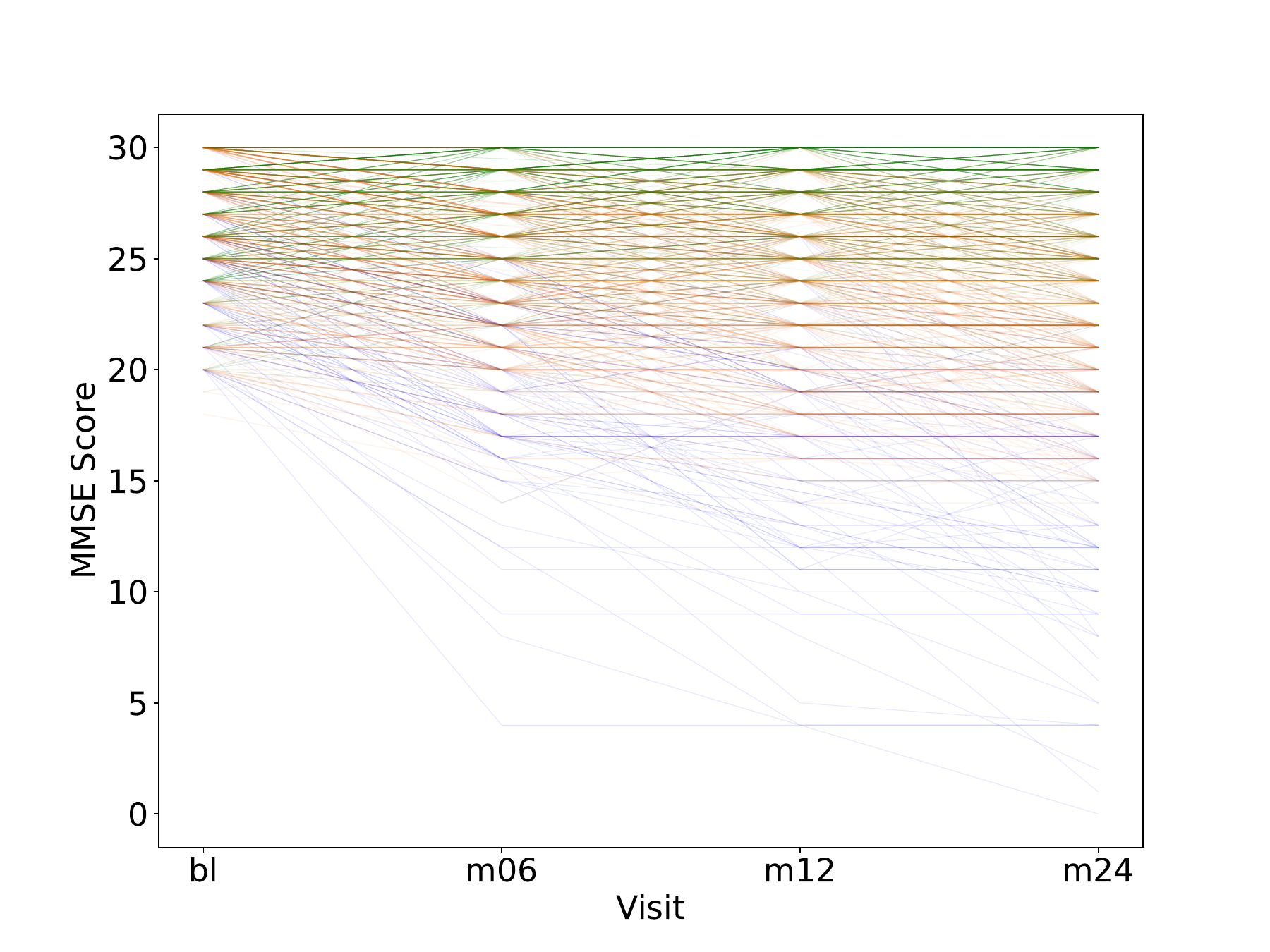}}%\hfill
	\caption{AD subtype clusters based on the decrease of MMSE at each visit. (a) each line represents the average score across patients for each cluster and the shadow represents one standard deviation. (b) individual lines per patient are plotted.}
	\label{fig:adni_per_clust}
\end{figure}

\section{Materials and Experiments}

\subsection{Dataset}
\label{sec:dataset}
The Alzheimer’s Disease Neuroimaging Initiative (ADNI) database  is a landmark dataset for the advancement of our understanding of Alzheimer's disease. ADNI is comprised by a wide range of data modalities including MRI and PET images, genetics, cognitive tests, CSF and blood biomarkers (for up-to-date information, see www.adni-info.org). Longitudinal data for all subjects was selected as up to two years of progression after disease onset due to very high missingness rate (percentage of data points missing across patients for a given time point) present for time points posterior to the two years.
% Data used in the preparation of this article were obtained from the Alzheimer's Disease Neuroimaging Initiative (ADNI) database (adni.loni.usc.edu). The ADNI was launched in 2003 as a public-private partnership, led by Principal Investigator Michael W. Weiner,MD. The primary goal of ADNI has been to test whether serial magnetic resonance imaging (MRI), positron emission tomography (PET), other biological markers, and clinical and neuropsychological assessment can be combined to measure the progression of mild cognitive impairment (MCI) and early Alzheimer's disease (AD).
% Moreover, following the subtyping steps described in section~\ref{sec:subtype def}
Subjects were clustered using k-means in 3 main groups based on their Mini-Mental State Examination (MMSE) scores as can be seen in Fig~\ref{fig:adni_per_clust}. These clusters match the cognitive decline rate of patients over time. The MMSE score at each visit (basline, 6 months, 12 months, and 24 months) was used to determine the cognitive decline for each patient. As each patient may have a different starting level at baseline, the baseline measurement is subtracted from each of the following time points; thus all patients start at 0. Then, using k-means clustering, using $k = 3$, slow, intermediate and fast cognitive decline groups are defined.
As seen in Table~\ref{tab:demographics} the raw average MMSE score at baseline is comparable across all groups with a steep decrease for the fast and intermediate groups at 24 months. The slow or otherwise stable group MMSE score at 24 months is comparable to the one at the initial stage. On the other hand, all 3 groups are age matched. Similarly, gender distributions across groups is maintained with male subjects representing approximately 60\% of the subjects for each group.

% \subsection{Data preprocessing}
\textbf{Data preprocessing:}
The data for the imaging and genotype modalities was processed previously to the tokenization process following best practices for the corresponding modality as described below.

\begin{itemize}
\item \textbf{Imaging}: The FreeSurfer image analysis suite (\url{http://surfer.nmr.mgh.harvard.edu/}) is used to conduct cortical reconstruction and volumetric segmentation. T1 weighted MRI scans are segmented based on the FreeSurfer atlas for cross-sectional processing, enabling group comparison at a specific time point~\cite{fischl2000measuring}. For each reconstructed cortical region, cortical volume, thickness average, thickness standard deviation, and surface area measurements are labeled by the 2010 Desikan-Killany atlas. The UCSF ADNI team conducted this process~\cite{hartig2014ucsf}, and more information can be found at \url{http://adni.loni.usc.edu}.

\item \textbf{Genotype}: The genotype variants were filtered using the intersection between the \textit{List of AD Loci and Genes with Genetic Evidence Compiled by ADSP Gene Verification Committee} and the most recent genome wide association study (GWAS) on AD~\cite{bellenguez_new_2022}. The odds Ratios, rare allele frequency and intergenic region binary trait were obtained from the most recent GWAS study with accession number (GCST90027158), accessed through the GWAS catalog~\cite{buniello_gwas_2019}. Furthermore, the genotype variants were processed for sample and variant quality controls using PLINK1.9~\cite{purcell_plink_2007}.

\item \textbf{Clinical}: The clinical assessment features correspond to 7 different cognitive metrics available through ADNI. These are Logical Memory - Delayed Recall (LDELTOTAL), Digit Symbol Substitution (DIGITSCOR), Trail Making Test B (TRABSCOR) and Rey Auditory Verbal Learning Test (RAVLT) scores: immediate, learning, forgetting and percent forgetting.
Values for imaging and clinical modalities were normalized in each train set before they were used as inputs to the network.
\end{itemize}

% \begin{figure*}[t]
% \centering
% \includegraphics[width = 0.5\textwidth,trim=40 25 40 25, clip =true]{figures/adni_plot_kmeansclust_MMSE_survival_plot_one_plot.pdf}
% \caption{AD subtypes clusters based on the progression of MMSE at each visit. Thick lines represent the average score across patients for each cluster, shadow represents one standard deviation and individual patients are plotted with transparency.}
% \label{fig:adni_per_clust}
% \end{figure*}

\subsection{Experimental design}
% Benchmarks: single vs multimodal approaches in neuro
% non-attention fusion mechanisms
All models were trained and evaluated in 10-fold stratified cross-testing, 5-fold stratified cross-validation for hyperparameter tuning. First the full dataset was split into 10 folds where 1 fold is left out for testing and the rest for training, this training set is then split into 5 folds to perform hyperparameter tuning. Through this framework, all data leakage is prevented. The best hyperparameters were determined on the validation set at each experimental run by selecting the best performing model. %Further details on hyperparameter tuning, training resources and settings can be found in the supplementary materials.
Predictions were evaluated using the area under the receiver operating curve (AUROC). A one-vs-one strategy is employed where the average AUROC of all possible pairwise combinations of classes is computed for a balanced metric. This is implemented using Sci-Kit Learn API~\cite{pedregosa_scikit-learn_2011}, which implements the method described in \cite{hand_simple_2001}.
The model was compared against the stage-wise deep learning intermediate fusion model introduced in~\cite{zhou_effective_2019} and several well-established traditional ML models - random forest (RF), support vector machine (SVM) with radial-basis function (RBF) kernel. Similarly, each of the branches of the model was used as comparison using a series of transformer encoder layers and MLP head for classification.
%AUROC, precision-recall curves and confusion matrices were used to evaluate Tri-COAT’s and competing models performances. Due to the imbalanced the precision-recall curves allow for a better assessment of the model. 

Tri-COAT consist of four transformer layers for each of the single modality encoders, with four attention heads per transformer layer. The tri-model co-attention process is done in a single-head attention mechanism. The classifying MLP has one hidden layer with 256 units. Embedding dimension of $k=256$ was used for all modalities. The model dimensions for the single-modality encoders were kept at 256 and all throughout as this combination achieved the best results on the validation set. The final mlp had the concatenated class tokens resulting from the tri-modal co-attention module and computed the output logits for each one of the three possible classes. Adam was used to optimize both Tri-COAT and the stage-wise mlp model using learning rates of 0.0001 for Tri-COAT and 0.0001 for the stage-wise fusion benchmarking model. All deep learning models were trained using cross-entropy loss. All deep learning models were implemented using PyTorch, while the RF and SVM models were implemented using scikit-learn. All deep learning models were trained for 100 epochs and the best checkpoint, meaning epoch with the highest AUROC on the validation set, was selected for model evaluation. The stage-wise deep learning fusion model had dimensions of 64 units for the single-modality layers, 32 for the second-stage and 16 for the final stage. The model dimensions were selected following the described hyperparamethers in \cite{zhou_effective_2019}. The SVM used an rbf kernel, regularization parameter C of 1. The random forest used gini as its criterion for leaf splitting, 100 trees and no maximum depth.

\begin{figure*}[t]
\centering
\subfloat[Clinical and Imaging]{\includegraphics[width = 0.33\textwidth,trim=40 25 40 25, clip =true]{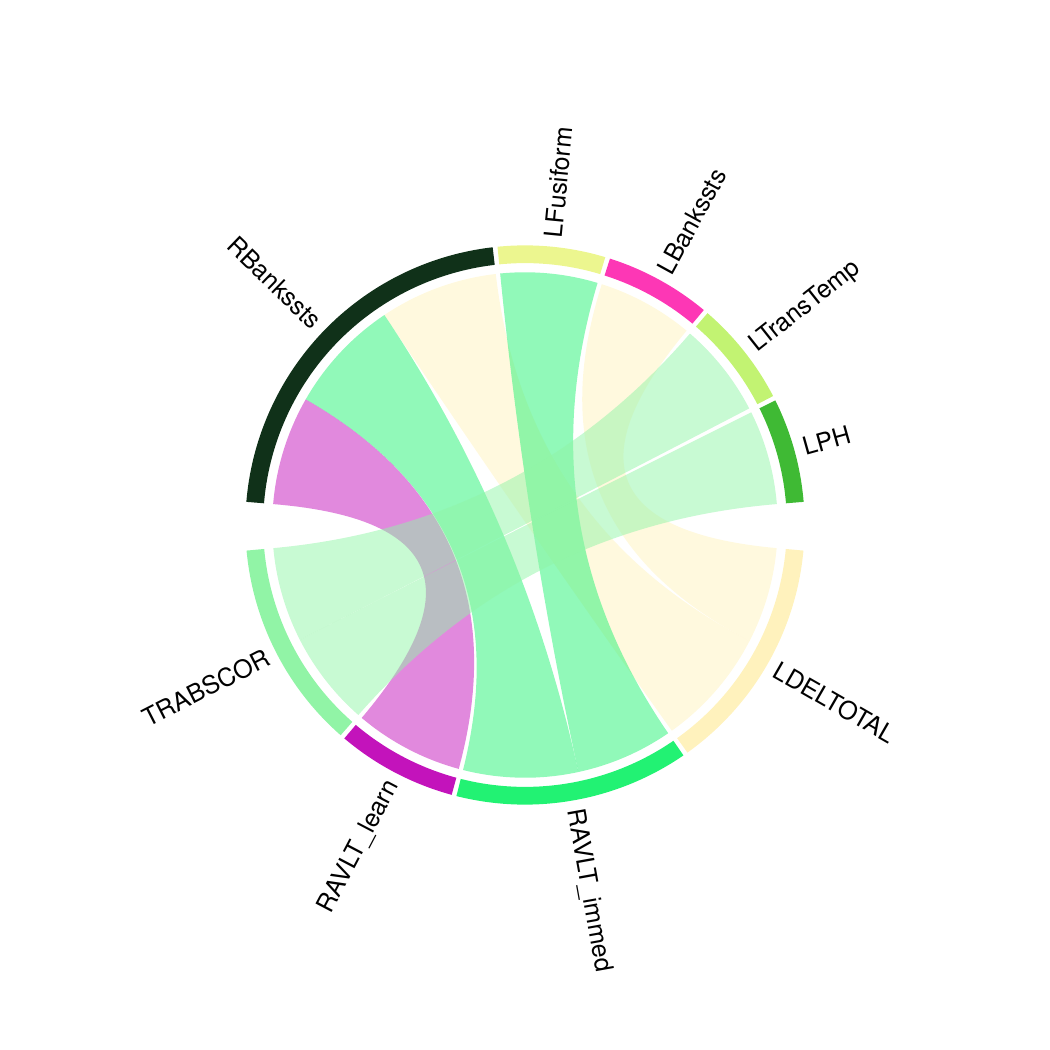}}%\hfill
\subfloat[Clinical and Genotype]{\includegraphics[width = 0.33\textwidth,trim=40 25 40 25, clip =true]{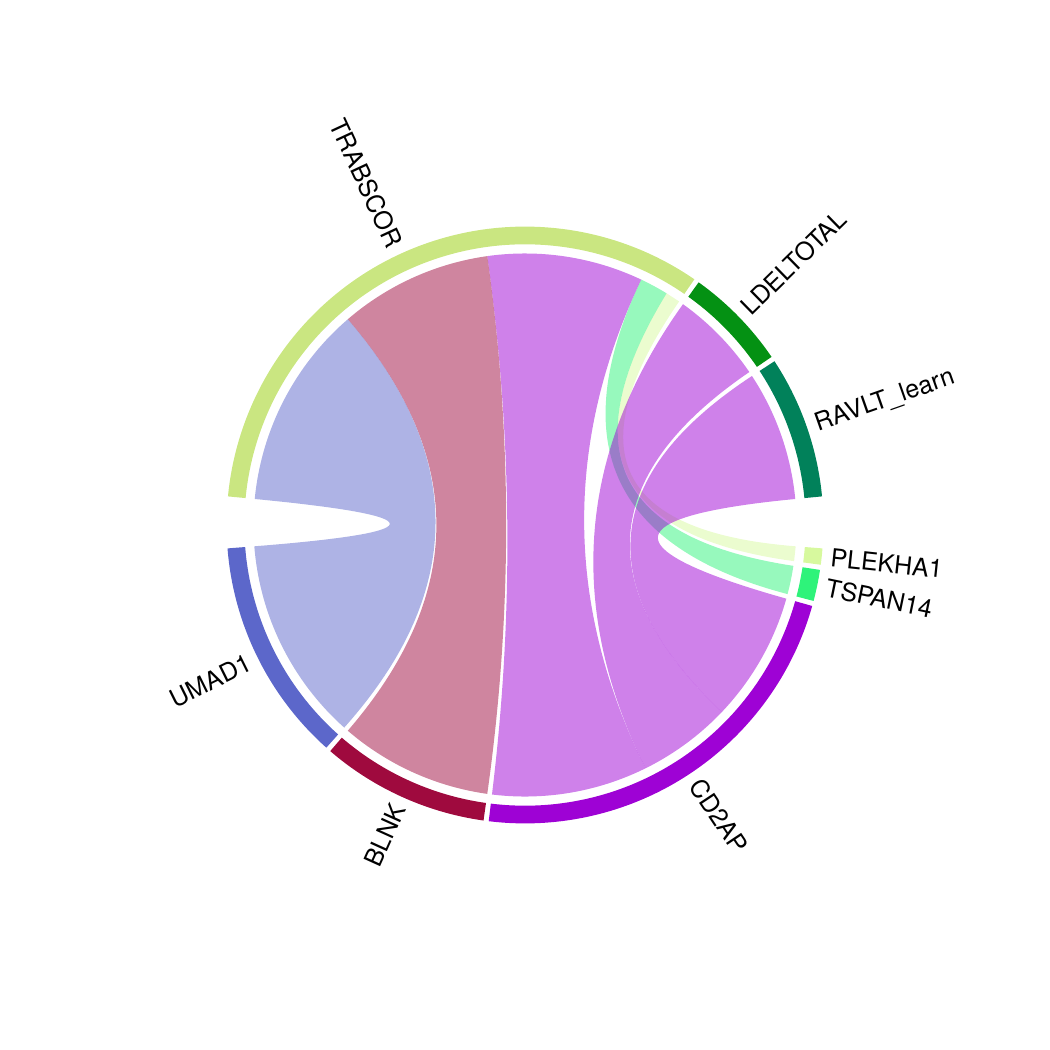}}%\hfill
\subfloat[Imaging and Genotype]{\includegraphics[width = 0.33\textwidth, trim=40 20 40 25, clip =true]{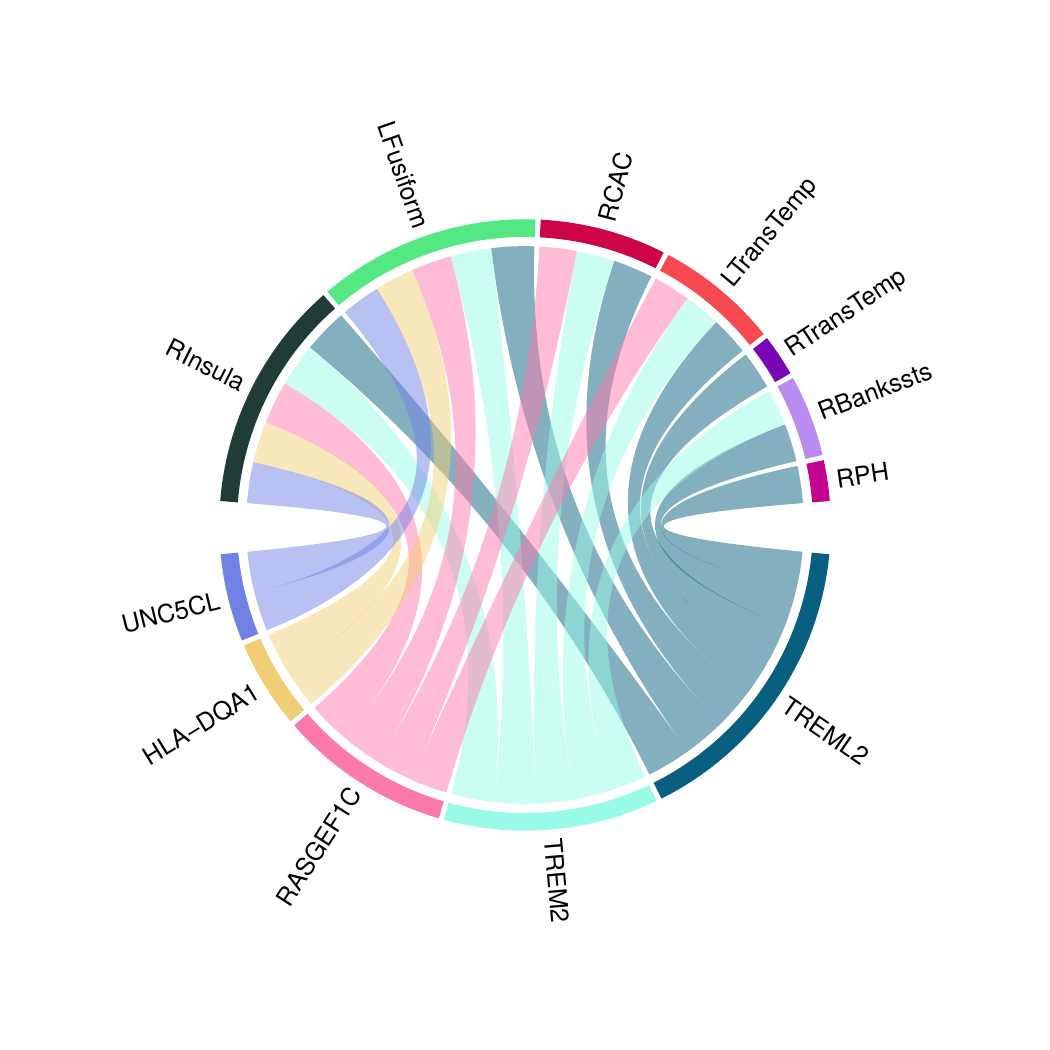}}%\hfill
\caption{AD key biomarker associations form learned co-attention.}
%Left: Clinical and Imaging. Center: Clinical and Genotype Right: Imaging and Genotype.}
\label{fig:adni_img_clin_attn}
\end{figure*}

\begin{table*}[t]
	\centering
	\caption{Mean AUROC $\pm$ SD of 10-fold cross-testing results. The proposed model significantly outperformed all the baseline models. The statistical significance was evaluated by paired $t$-test with $\alpha=0.005$, except for the entry where $\alpha=0.05$.
		%Values with significant difference ($\alpha=0.05$) using paired t-test denoted with '\mbox{*}' and '\mbox{**}' for $\alpha=0.005$.} %\py{The p-values look significant (0.010), don't they?} ($\alpha=0.05$).
}
\label{tab:results}
%\begin{tabular}{l|c|c|c|c}
\begin{tabular}{p{80pt} | p{50pt} | p{50pt}| p{50pt} | p{50pt}}
	\toprule
	% \begin{tabular}{cccccc}
		% \toprule
		%\textbf{Model} & \textbf{Mean ROC AUC $\pm$ SD} & \textbf{Mean Precision-Recall AUC $\pm$ SD}& \textbf{P-value}\\
		% \multicolumn{1}{c}{Model} & \multicolumn{2}{c}{Mean ROC AUC $\pm$ SD} & \multicolumn{2}{c}{Mean Precision-Recall AUC $\pm$ SD} & \multicolumn{2}{c}{P-value} \\
		% &\textbf{Model} & \textbf{Mean AUROC $\pm$ SD}\\
		% \midrule
		
		% & RF & 0.705 $\pm$ 0.036 \mbox{*}\\
		% & SVM           & 0.684 $\pm$ 0.0.047 \mbox{**}\\
		% & Our model   &\textbf{0.733 $\pm$ 0.070} \\
		
		Method & Full          & Imaging         & Genetics        & Clinical        \\ 
		% & \multicolumn{4}{c}{Mean AUROC $\pm$ SD}                              \\
		\midrule
		%SVM   & \textit{0.705 ± 0.036}\mbox{*} & 0.669   ± 0.060\mbox{**} & 0.525   ± 0.034\mbox{**} & 0.639  ± 0.078\mbox{**}  \\
		%RF    & 0.684 ± 0.0477\mbox{**} & 0.67   ± 0.052\mbox{**} & 0.505   ± 0.031\mbox{**} & 0.659  ± 0.087\mbox{**}  \\
		%Tri-COAT  & \textbf{0.734 ± 0.076} & 0.648   ± 0.056\mbox{**} & 0.539 ± 0.084\mbox{**} & 0.697   ± 0.063\mbox{**} \\ 
		SVM   & \textit{0.705 $\pm$ 0.036} & 0.669 $\pm$ 0.060 & 0.525 $\pm$ 0.034 & 0.639 $\pm$ 0.078  \\
		RF    & 0.684 $\pm$ 0.048 & 0.677 $\pm$ 0.052 & 0.505 $\pm$ 0.031 & 0.659 $\pm$ 0.087  \\
		Stage-wise fusion\cite{zhou_effective_2019} & 0.641 $\pm$ 0.017 & 0.557 $\pm$ 0.096 & 0.562 $\pm$ 0.078 & 0.655 $\pm$ 0.057  \\
		Tri-COAT  & \textbf{0.734 $\pm$ 0.076} & 0.648 $\pm$ 0.056 & 0.539 $\pm$ 0.084 & 0.697 $\pm$ 0.063 \\ 
		
		\bottomrule
	\end{tabular}
\end{table*}

\begin{table}[t]
\centering
\caption{Mean AUROC $\pm$ SD of 10-fold cross-testing results. The proposed model significantly outperformed all the baseline models. The statistical significance was evaluated by paired $t$-test with $\alpha=0.005$, except for the entry where $\alpha=0.05$.
%Values with significant difference ($\alpha=0.05$) using paired t-test denoted with '\mbox{*}' and '\mbox{**}' for $\alpha=0.005$.} 
}
\label{tab:fusion}
\begin{tabular}{p{48pt} | p{50pt}}
\toprule
Method & AUROC        \\ 
\midrule

Early   & \textit{0.571 $\pm$ 0.053} \\
Late    & 0.604 $\pm$ 0.048  \\
Tri-COAT  & \textbf{0.734 $\pm$ 0.076} \\ 

\bottomrule
\end{tabular}
\end{table}

\section{Results and Discussion}
\subsection{Clustering, label definition}
Following the literature, the number of clusters was set to three main groups. The MMSE score was used as an indicator of mental decline. Based on the speed of the progression of the mental decline over a period of two years, three groups were defined, i.e., fast, intermediate and slow progressing subtypes. K-means clustering was used to assign each subject to one of the groups.
% Moreover, the “elbow method” was used to further confirm the ideal number of clusters.
Through this process labels were defined for all subjects.
Only baseline data was used as input to Tri-COAT and all competing models. Based on the baseline data Tri-COAT was able to effectively classify the subjects into their corresponding subtypes. 
\subsection{AD subtype classification}
As seen in Table~\ref{tab:results}, Tri-COAT outperformed the single modality ablations and baseline models, achieving an average AUROC of 0.733 $\pm$ 0.070 across all test sets in the 10-fold cross-testing framework.
For the single-modaltiy ablation studies, each modality was used independently to classify the AD subtype. For Tri-COAT, a single-modality transformer encoder backbone and MLP head were used. For the stage-wise fusion model, it was adapted to MLPs using the same number of hidden layers as the first plus last stage of the multimodal version. For SVM and RF, no variations were required. Each modality was evaluated using the same 10-fold cross testing - 5 fold-cross validation hyperparameter tuning framework.
Moreover, the single modality ablation models outperformed their baseline counterparts for the clinical and genetics modalities. In contrast to this the baselines achieved better performances for the imaging derived traits.
For the three modalities, clinical achieves the best classification AUROC, followed by imaging and genetics. This is expected as biologically, the same order follows for the closest relation between the observed phenotype and the mechanisms behind it. Clinical (cognitive) assessments are the closest to the MMSE metric, followed by imaging (changes in the brain morphology) which is directly related to the observed phenotype and genetics being the farthest apart from the expressed symptoms.
Both comparative models and Tri-COAT achieve higher performances in their multimodal configuration compared to single modalities, agreeing with previous literature regarding multimodal approaches for classification of AD and related disorders. 
% \py{How does Tri-COAT acquire the single modality performance in Table 2?}

Furthermore, as seen in Table~\ref{tab:fusion}, Tri-COAT outperformed variations of itself using alternative fusion strategies. The early fusion model considerably underperforms achieving an AUROC of 0.571 $\pm$ 0.053 because of its limited capabilities to simultaneous encode highly heterogeneous data with distinct biological level relationships to the outcome. Similarly, the late fusion model underperforms as it is limited to joining the predictions from each branch and cannot learn the relationships between the different modalities in the latent space.

\begin{figure*}[t]
\centering
\includegraphics[width = \textwidth,trim=0 0 0 0, clip =true]{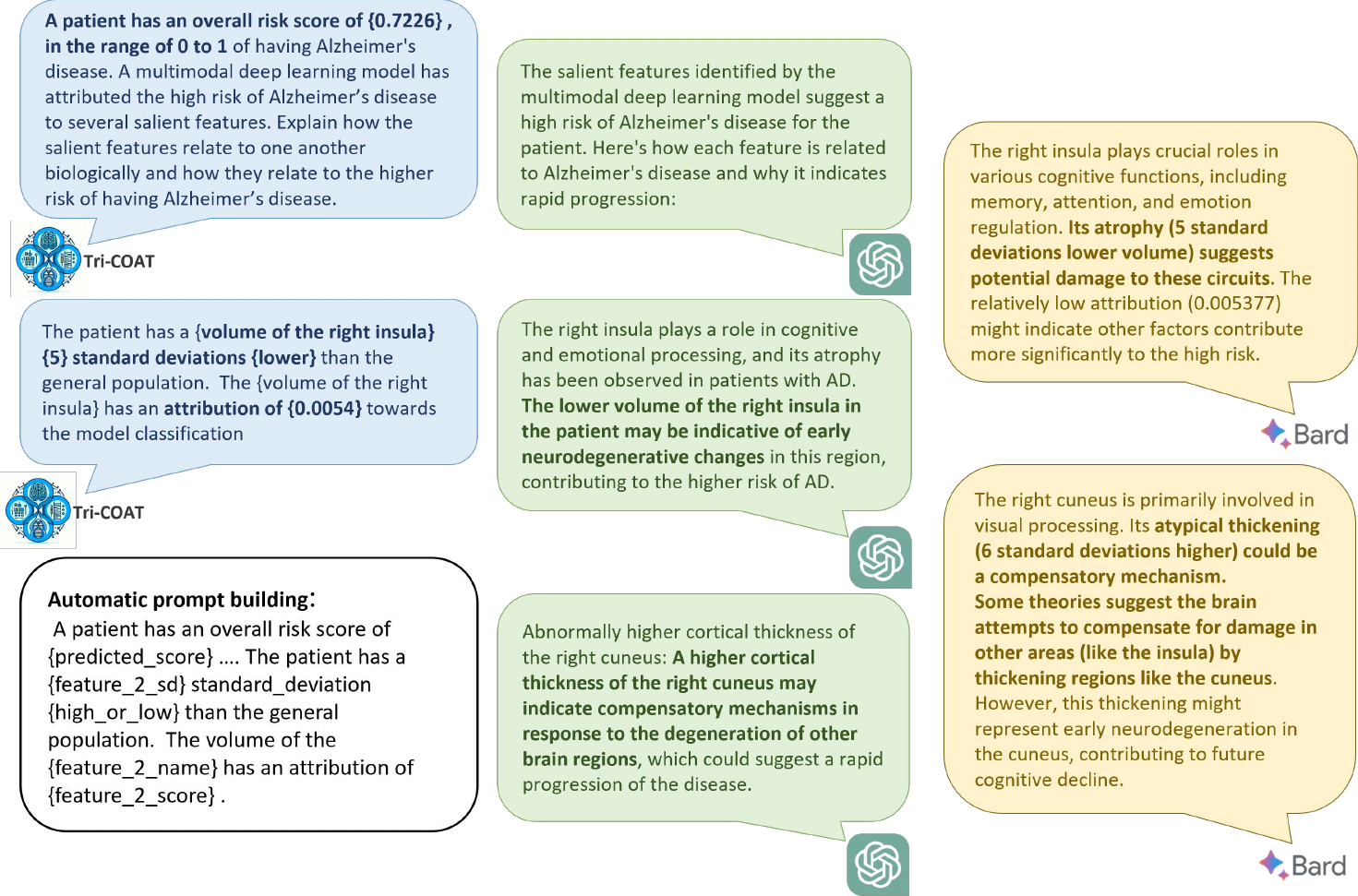}%\hfill
% \subfloat{\includegraphics[width = 0.5\textwidth,trim=100 20 100 0, clip =true]{figures/chat_gpt_1.pdf}}%\hfill
% \subfloat{\includegraphics[width = 0.5\textwidth, trim=100 20 100 0, clip =true]{figures/chat_gpt_2.pdf}}%\hfill
\caption{Tri-COAT interpretability through ChatGPT. Prompts were built based on Tri-COAT's prediction and integrated gradients feature attributions for each patient. Example portions of the prompts are showing in the blue text bubbles and ChatGPT answers are shown in the green text bubbles.}
\label{fig:chat_gpt}
\end{figure*}

\subsection{Biomarker Associations Learned by Co-attention}
One of the key advantages of Tri-COAT compared to the baseline models and traditional deep learning approaches is the ability to learn insights into the cross-modal feature associations. In order to explore the learned relationships, the model with highest test AUROC from the evaluation framework was selected for attention visualization in which the learned attention scores were averaged across all test subjects. Chord plots were drawn using the circlize R library~\cite{gu_circlize_2014} to visualize the cross-modal attention.
As seen in Figure~\ref{fig:adni_img_clin_attn}, Tri-COAT identified key associations between the Trails making test B score (TRABSCORE) in clinical-imaging and clinical-genetics. This score tests for cognitive ability of the patient for working memory and secondarily task-switching ability~\cite{terada_trail_2013}. Clinical literature shows strong correlation between gyri structures - temporal gyrus and Parahippocampal gyrus (LTransTemp, LPH) and TRABSCORE~\cite{matias-guiu_neural_2017}. Similarly, for the clinical-genotype association, TRABSCORE was found to be associated with the CD2AP gene which has been clinically identified as a driver for the AD hallmark - neurofibrillary tangles (NFT) in the temporal gyrus region~\cite{camacho_association_2021}. This is a very exciting finding for our network as it establishes a putative relationship between genetics (CD2AP gene), brain ROIs (temporal gyrus) and clinical symptoms (TRABSCORE).
% Furthermore, the relationship between XXX and XXX has been shown to be essential for AD development as it is part of the XXX biological pathway.

% \begin{figure*}[t]
% \centering
% \includegraphics[width = 0.5\textwidth,trim=0 0 0 0, clip =true]{figures/confusion_mat.pdf}%\hfill
% \caption{AD }
% \label{fig:confusion_mat}
% \end{figure*}

\subsection{Tri-COAT assessments explained through ChatGPT and Bard}

In this section, we investigate using large language models, namely ChatGPT (GPT-3.5 March 23, 2023 version)~\cite{openai_gpt-4_2023} and Bard(Accessed Jan 30th 2024)~\cite{google_bard_alzheimers_nodate}, to interpret the results of multimodal deep learning models. %Deep learning models are traditionally referred as "black-box" approaches as they have limited interpretability.
In order to enhance the interpretability of Tri-COAT, descriptions of each patient case (“prompts”) are built using the top-10 features with their corresponding standard deviation and ChatGPT is employed to interpret said prompts. First, the attribution for each input feature were calculated using integrated gradients~\cite{sundararajan_axiomatic_2017}. Next, based on the absolute contribution, the most salient features for each patient were determined. Finally, a text prompt for each patient was built using the most salient features. The prompt describes the quantitative attribution of the top-10 salient features and how many standard deviations the patient's measurements are from the whole sample distribution. ChatGPT provided very informative answers explaining the biological relevance of each salient feature towards AD development. A short example can be seen in \ref{fig:chat_gpt}. 
% Further extended examples can be seen in supplementary data. 
%

To provide more detail, we built a prompt for each patient based on the top-weighted features obtained from integrated gradients. The prompts were structured as a question and a case describing the patient biomarker measurements. The question asks to explain the model prediction and the case contains the selected features’ attribution and corresponding standard deviation. The attribution score obtained for each feature through integrated gradients describes how relevant each feature is toward classifying each patient.  The input features are ranked using the attribution scores and the top 10 features are selected for each subject. In order to describe how each subject differs from the rest of the group based on the analyzed biomarkers, the standard deviation for every feature is calculated. A template was built to be filled out with the patient specific data. The template was iteratively refined to improve the LLM output and its readability. The template followed a 11 short descriptions structure, the first being the patient risk score assessment plus one description per salient feature. For the patient risk assessment model, it started by introducing the patient risk and the predictive model as seen in Fig.~\ref{fig:chat_gpt}. Next, 10 short descriptions of the top-10 attributions, including slots for the salient feature name, the corresponding standard deviation of the patient measurement and if this represents a higher or lower value compared to the population.
 % We used ChatGPT  (GPT-3.5 March 23, 2023 version) and Bard (Accessed Jan 30th 2024) to generate interpretable prediction reports per subject.

The reports generated by ChatGPT and Bard includes clear descriptions for each feature of interest and their relationship toward AD development and how jointly they impact the deep learning model’s assessment. This approach provides not only high explainability to the model assessment but also provides key insights into cross-modal feature associations supported by known biological mechanisms. 
 % The prompts were analyzed by a large language model, namely ChatGPT (GPT-3.5 March 23, 2023 version). A final report for each subject that interprets the model AD assessment was obtained as the output from ChatGPT.
%ChatGPT provided very informative answers explaining the biological relevance of each salient feature towards AD development. The prompt describes the numerical contribution of each feature to the model classification and how different the patient-measured trait is from the rest of the group via the standard deviation. We used the top 10 features in our experiments. In turn, ChatGPT describes the biological significance of each feature and how it relates to AD development. ChatGPT is able to explain the classification based on the measured trait of each patient.
For example, ChatGPT describes the biological functions of the right insula and its connection with AD. Next, it explains that the significantly reduced volume in this brain structure suggests a loss of function and a rapidly progressing AD~\cite{zhu_brain_2022}.  Similarly, it is exciting to see that even when the measured trait deviates from the group in the opposite direction as expected for an AD patient, ChatGPT identifies the value as unexpected and provides an explanation for it. For example, for the abnormally higher cortical thickness of the right cuneus, it explains that this could indicate a compensatory mechanism~\cite{pettigrew_cognitive_2017}. We saw similar responses from both ChatGPT and Bard, which results matched findings known in clinical literature. It is however important to note that LLMs have been found to hallucinate~\cite{huang_survey_2023}. While there are promising approaches to mitigate hallucinations; this is a key limitation of current LLMs towards implementation in clinical settings. 

\section{Conclusions}
AD is the most prevalent neurodegenerative disease, and all current treatments are limited to slowing disease progression; thus making early diagnosis essential. Furthermore, there are multiple subtypes with different rates of cognitive decline. In order to move closer to personalized medicine, it is essential to have a better understanding of the heterogeneity surrounding the disease development. However, early subtyping is a very challenging task. Our proposed model was able to effectively classify AD patients into three main subtypes using prodromal factors measured at baseline. 
Moreover, the model was able to identify multiple putative cross-modal biomarker networks. Nevertheless, the generalizability of the learned features on other datasets remains to be tested. We demonstrated that multimodal deep learning models can achieve increased interpretability through large language models and attribution methods; which, in turn, can lead in future work to personalized medicine applications of deep learning models in clinical settings.  Similarly, in future work we propose to extend this framework to other heterogeneous neurodegenerative diseases such as Parkinson's disease.

\section{Acknowledgements}
Data collection and sharing for this project was funded by the Alzheimer's Disease Neuroimaging Initiative (ADNI) (National Institutes of Health Grant U01 AG024904) and DOD ADNI (Department of Defense award number W81XWH-12-2-0012). ADNI is funded by the National Institute on Aging, the National Institute of Biomedical Imaging and Bioengineering, and through generous contributions from the following: AbbVie, Alzheimer's Association; Alzheimer's Drug Discovery Foundation; Araclon Biotech; BioClinica, Inc.; Biogen; Bristol-Myers Squibb Company; CereSpir, Inc.; Cogstate; Eisai Inc.; Elan Pharmaceuticals, Inc.; Eli Lilly and Company; EuroImmun; F. Hoffmann-La Roche Ltd and its affiliated company Genentech, Inc.; Fujirebio; GE Healthcare; IXICO Ltd.;Janssen Alzheimer Immunotherapy Research \& Development, LLC.; Johnson \& Johnson Pharmaceutical Research \& Development LLC.; Lumosity; Lundbeck; Merck \& Co., Inc.;Meso Scale Diagnostics, LLC.; NeuroRx Research; Neurotrack Technologies; Novartis Pharmaceuticals Corporation; Pfizer Inc.; Piramal Imaging; Servier; Takeda Pharmaceutical Company; and Transition Therapeutics. The Canadian Institutes of Health Research is providing funds to support ADNI clinical sites in Canada. Private sector contributions are facilitated by the Foundation for the National Institutes of Health (www.fnih.org). The grantee organization is the Northern California Institute for Research and Education, and the study is coordinated by the Alzheimer's Therapeutic Research Institute at the University of Southern California. ADNI data are disseminated by the Laboratory for Neuro Imaging at the University of Southern California

% \section{References}
\bibliographystyle{IEEEtran}
\bibliography{main_article}

\end{document}